\documentclass{article}

\PassOptionsToPackage{numbers, compress}{natbib}

\usepackage[final]{neurips_2021}

\usepackage[utf8]{inputenc} 
\usepackage[T1]{fontenc}    
\usepackage{hyperref}       
\usepackage{url}            
\usepackage{booktabs}       
\usepackage{amsfonts}       
\usepackage{nicefrac}       
\usepackage{microtype}      
\usepackage{xcolor}         
\usepackage{dsfont}         
\usepackage{paralist}		
\usepackage{wrapfig}	
\usepackage{tikz}	
\usetikzlibrary{arrows,positioning,arrows.meta}
\usepackage [fleqn] {amsmath}	

\usepackage{graphicx} 
\usepackage{caption}        
\usepackage{setspace}       
\usepackage{array}   		
\usepackage{algorithm}
\usepackage[noend]{algpseudocode}	
\usepackage{amssymb}        

\newcolumntype{R}{>{\raggedleft}p{19.7mm}}

\newcommand{\HSIC}{\mathsf{HSIC}}
\newcommand{\mse}{\mathsf{MSE}}
\newcommand{\var}{\mathsf{Var}}
\newcommand{\loss}{\mathsf{loss}}
\newcommand{\lossVAEy}{\mathsf{loss}_{\mathsf{VAE},y}}
\newcommand{\lossVAEx}{\mathsf{loss}_{\mathsf{VAE},x}}
\newcommand{\net}{\mathsf{net}}
\newcommand{\DKL}{\mathsf{D}_\mathsf{KL}}

\newcommand{\I}{\mathcal{I}}

\title{Encoding Causal Macrovariables}

\author{%
  Benedikt Höltgen\thanks{Work done while a master student at the MCMP/LMU Munich.} \\
  External collaborator to OATML Group\\
  Department of Computer Science\\
  University of Oxford\\
  Oxford, United Kingdom \\
  \texttt{benedikt.hoeltgen@mailbox.org} \\
}

\newcommand{\RR}{\mathds{R}}

\newcommand{\obar}[1]{\mkern 1.5mu\overline{\mkern-1.5mu#1\mkern-1.5mu}\mkern 1.5mu}

\newcolumntype{L}{>{$}l<{$}} 
\newcolumntype{C}{>{$}c<{$}} 

\begin{document}

\maketitle

\begin{abstract} 
In many scientific disciplines, coarse-grained causal models are used to explain and predict the dynamics of more fine-grained systems.
Naturally, such models require appropriate macrovariables.
Automated procedures to detect suitable variables would be useful to leverage increasingly available high-dimensional observational datasets.
This work introduces a novel algorithmic approach that is inspired by a new characterisation of causal macrovariables as information bottlenecks between microstates.
Its general form can be adapted to address individual needs of different scientific goals.
After a further transformation step, the causal relationships between learned variables can be investigated through additive noise models.
Experiments on both simulated data and on a real climate dataset are reported.
In a synthetic dataset, the algorithm robustly detects the ground-truth variables and correctly infers the causal relationships between them.
In a real climate dataset, the algorithm robustly detects two variables that correspond to the two known variations of the El Ni\~{n}o phenomenon.
\end{abstract}

\section{Introduction} 

With graphical and structural causal models becoming increasingly popular, both scientists and philosophers are starting to look more closely at the main ingredient to these models: causal variables.
Finding such causal variables has long been a neglected area of research that has only recently started to motivate a growing literature in both machine learning (\cite{Chalupka2015, Chalupka2016a, Chalupka2016b}) and philosophy of science \cite{Woodward2016, Eberhardt2016}.
Especially in the machine learning community, there are now calls for further research into causal representation learning \cite{Scholkopf2019, Scholkopf2021} and variable construction in particular \cite{Eberhardt2017}.
Beyond artificial intelligence research, this is particularly relevant for scientific disciplines -- such as climate science, neuroscience, or economics -- in which higher-level models need to be constructed based on high-dimensional observational data.
The two central challenges are the identification of suitable macrovariables and the inference of causal relationships from purely observational data.

The approach proposed in this work is based on a novel characterisation of causal macrovariables as information bottlenecks.
Building on the information bottleneck framework, we show how neuron activations in artificial neural networks can be interpreted and optimised as coarse-grained causal variables over high-dimensional data.
To this end, we introduce a novel neural network structure loosely based on Variational Autoencoders, which we call the Causal Autoencoder.
It can be applied to settings where two high-dimensional datasets are available.
This framework allows to establish a connection between the often separately studied problems of causal inference (where both cause and effect variables are investigated) and learning disentangled representation (where only one dataset is given).
With the novel approach, the causal relationships between detected macrovariables can be investigated through additive noise models, after applying an additional transformation step.
The methodology is tested on both simulated and natural data.
For the simulated dataset, the ground-truth generative model is recovered, including the direction of causality.
For the natural climate dataset, sensible macrovariables are detected that are in line with corresponding domain knowledge.

\section{Background: Causal macrovariable detection} 

\subsection{Causal macrovariables} 
\label{ssec:causal_macro}

On a strict notion of micro- and macrovariables, they stand in a deterministic functional relationship: there needs to be some deterministic function between a (often high-dimensional) microvariable space and a macrovariable space.
This function assigns a macrovariable state to each microvariable state, the former 'supervene' on the latter.
It should be noted that micro- and macrovariable are relative notions: While temperature is a macrovariable in relation to the kinetic energy of molecules, it is a microvariable in the context of large-scale climate models with hundreds of temperature measurements.
In general, different scientific goals often require different scientific ontologies for the same system \cite{Danks2015, Potochnik2017}.

A good way to think about causal structures in the world is as relationships between patterns that supervene on microvariable states \cite{Andersen2017, Potochnik2017}.
There need, however, not be a unique causal structure within a given system: different ways of carving up the system into patterns can often yield a variety of causal structures within the system \cite{Potochnik2017}.
Furthermore, as \citet{Spirtes2007} and \citet{Eberhardt2016} show, even one specific causal structure ``can be equivalently described by two different sets of variables that stand in a non-trivial translation-relation to each other'' \cite{Eberhardt2016}.
In general, there is no a uniquely 'correct' choice of appropriate variables for the representation of causal systems;
this is sometimes explicitly acknowledged in the machine learning literature \cite{WeichwaldPhD}.

However, it does not mean that any model is as good as the other.
In fact, coming up with sensible scientific ontologies is one of the main tasks that scientists are concerned with.
Deciding between them can be guided by different criteria whose importance varies with context and goal.
James \citet{Woodward2016} has recently proposed a tentative list of criteria for causal variable selection, although conceding that eventually it might turn out ``that there is nothing systematic to say about this issue'' (p. 1048); among his criteria, the following four are particularly interesting for the present work (p. 1054f):
\begin{enumerate}[a)] 
\itemsep0em
\item ``there is a clear answer to the question of what would happen if they were to be manipulated or intervened
on''
\item they ``lead to causal representations that are relatively sparse''
\item they ``exhibit strong correlations between cause and effect''
\item the relationships between them ``continue to hold under changes in background conditions''
\end{enumerate}

\subsection{Problem setup} 

For this work we are considering a very general setup where two high-dimensional and dependent 
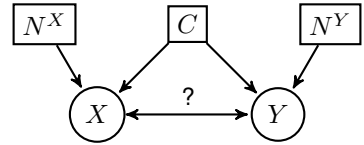
\begin{wrapfigure}{r}{0.37\textwidth}
\vspace{-2pt}
\centering
\begin{tikzpicture}[->,>=stealth',shorten >=1pt,auto,node distance=1.7cm, thick, 
			main node/.style={circle,draw},
                   	mult node/.style={rectangle,draw}]
  \node[mult node] (1) {$N^X$};
  \node[main node] (2) at (.7,-1.2) {$X$};
  \node[mult node] (3) [above right of=2] {$C$};
  \node[main node] (4) [below right of=3] {$Y$};
  \node[mult node] (5) at (3.8, 0) {$N^Y$};

  \path[every node/.style={font=\sffamily\small}]
    (1) edge node {} (2)
    (2) edge [<->]node [style= <->] {?} (4)
    (3) edge node {} (2)
    (3) edge node {} (4)
    (5) edge node {} (4);
\end{tikzpicture}
\caption{Schematic causal diagram of the setup with variables $X$ and $Y$, noise, and common causes. It is assumed that only $X$ and $Y$ are observed and that the graph is acyclic.}\label{Setup}
\vspace{-40pt}
\end{wrapfigure}
variables $X$ and $Y$ are given.
$X$ and $Y$ might have common causes $C$ and there might be a 
one-directional causal path either from $X$ to $Y$ or vice versa.\footnote{Throughout this work, uppercase letters denote multidimensional variables while lowercase letters denote one-dimensional ones.}
As the setup is not restricted to deterministic cases, we also allow that $X$ and $Y$ can be influenced by individual and independent noise $N^X$ and $N^Y$, respectively.
This leads to the setup depicted in fig.~\ref{Setup} and one of the following pairs of structural assignments:

$\begin{aligned}
\ &X := \alpha(N^X, C)\ & &Y = \beta(N^Y, C, X)\ & (X \to Y)\\
\ &X := \alpha(N^X, C, Y) & &Y = \beta(N^Y, C) & (X \leftarrow Y)\\
\ &X := \alpha(N^X, C) & &Y = \beta(N^Y, C) & (X \nleftrightarrow Y)
\end{aligned}$

\subsection{Related work} 

A task that is closely related to one addressed here is learning disentangled representations.
The goal of this area of research is to find representations that correspond to the (perhaps causal) factors of variation \cite{Bengio2013, Locatello2019}.
Various approaches to this have been based on Variational Autoencoders (VAEs), thus encoding samples into a noisy bottleneck layer and decoding it to 'predict' the input again.
Although the concept of mutual information (MI) is a cornerstone of many of these approaches, the relevance of MI for unsupervised representation learning algorithms is still unclear \cite{Tschannen2020}.
A central difference in the present approach is the aim of encoding \emph{two} high-dimensional datasets into bottlenecks that stand in causal relation to each other.

In other work, machine learning researchers have started to investigate causal relationships between neuron activations and image classification outputs \cite{Lopez-Paz2017}.
However, this type of causal relationship concerns the algorithm’s classification mechanism rather than dependencies in the data.
The authors investigate claims like ``the presence of cars cause the presence of wheels'' rather than actual causal mechanisms in the world.
While they also use neuron activations, their activations do not provide information bottlenecks between two datasets.

The closest work to the present one, focusing on the same setup, is the 'causal feature learning' approach developed by Krzysztof Chalupka and colleagues \cite{Chalupka2015, Chalupka2016a}.
Their aim is to find categorical variables representing different causal macrostates in each of the two datasets $X$ and $Y$.
The approach assumes that the causal direction is known beforehand.
The key idea is that $X$-microstates belong to the same causal $X$-macrostate iff, when the result of an intervention, they induce the same probability distribution over $Y$-microstates and analogous for $Y$-microstates.
In \cite{Chalupka2016b}, they report results from applying their algorithm scheme to climate data, interpreting them as an ``unsupervised discovery of El Ni\~{n}o''.
The resulting causal macrovariables are categorical, which are strictly less informative than continuous ones.
Another limitation, which they concede in the context of the mentioned climate data, is that without (perhaps infeasible) real climate experiments or ``large-scale climate experiments with detailed climate models'' \cite{Chalupka2016b}, no causal claims can be justified.
While their work provides a potentially very fruitful avenue, we will in the following suggest a novel approach aiming to overcome these limitations by detecting continuous macrovariables.

\section{Method}

\subsection{Causal macrovariables as information bottlenecks} 
\label{ssec:bottleneck}

\begin{wrapfigure}{r}{0.43\textwidth}
\centering
\begin{tikzpicture}[->,>=stealth',shorten >=1pt,auto,node distance=1.5cm, thick, 
			main node/.style={circle,draw,minimum size=0.8cm}]
  \node[main node] (nx1) {$n^x_1$};
  \node[main node] (x1) [right of=nx1] {$x_1$};
  \node[main node] (x2) [below of=x1,yshift=5mm] {$x_2$};
  \node[main node] (c) [above right of=x1] {$c_1$};
  \node[main node] (y1) [below right of=c] {$y_1$};
  \node[main node] (y2) [below of=y1,yshift=5mm] {$y_2$};
  \node[main node] (ny1) [right of=y1] {$n^y_1$};
  \node[main node] (ny2) [right of=y2] {$n^y_2$};

  \path[every node/.style={font=\sffamily\small}]
    (nx1) edge node {} (x1)
    (c) edge node {} (x1)
    (c) edge node {} (y1)
    (x2) edge node {} (y2)
    (ny1) edge node {} (y1)
    (ny2) edge node {} (y2);
\end{tikzpicture}
\caption{Example of a simple causal diagram with the macrovariables $x_1$, $y_1$, $x_2$, and $y_2$, where the former two have a common cause and the third causes the fourth.}\label{two_macrovar}
\end{wrapfigure}
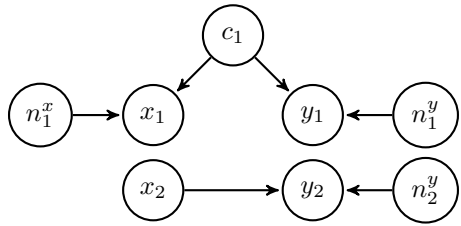

The starting point for our approach to automated causal macrovariable detection is the insight that all dependencies are due to causation.
This was famously formulated by \citet{Reichenbach1956} in his principle of the common cause:
If events -- or, rather, random variables -- A and B are correlated -- or, rather, dependent --, then either A caused B, B caused A, or A and B are both effects of a shared common cause.
This also implies that causal macrovariables provide an information bottleneck (sometimes also called sufficient statistics) between the microvariables.
To illustrate this, assume that the simple causal diagram of fig.~\ref{two_macrovar}, including the causal macrovariables $x_1$, $y_1$, $x_2$, and $y_2$, exhaustively represents the causal structure of the system depicted in fig.~\ref{Setup}.
Now, by exhausting all causal connections between $X$ and $Y$, the principle of the common cause implies that they also exhaust all mutual information shared by $X$ and $Y$.
In formal terms, $\I(\obar{X};Y) \geq \I(X;Y)$ and $\I(\obar{Y};X) \geq \I(Y;X)$ for $\obar{X} := x_1, x_2$ and $\obar{Y} := y_1, y_2$, where $\I(\cdot\ ; \cdot\cdot)$ denotes the mutual information.
As the macrovariables are functions of the microstates, they also cannot contain more information than the respective micro description.
This leads to the equalities $\I(\obar{X};Y) = \I(X;Y)$ and $\I(\obar{Y};X) = \I(Y;X)$.\footnote{Note that the example of fig.~\ref{two_macrovar} has a particularly simple structure.
If there was a pair of variables which both stand in a direct causal relationship and have common causes, this would not affect the information theoretic description given above, as the variables would still contain all the shared information. The causal inference techniques discussed below can, however, only handle cases with a simple structure.}

Recall that we want to find representations $f_i$ and $g_i$ yielding macrovariables $x_i = f_i(X)$ and $y_i = g_i(Y)$ which ideally capture all and only the \emph{mutual} information between $X$ and $Y$.
We noted earlier that abstracting to higher-level descriptions generally comes with a loss of information.
Precisely for this trade-off between compressing a signal and ``preserving the relevant information about another variable'', \citet{Tishby1999} developed the information bottleneck (IB) framework.
In this framework, the ``optimal assignment'' of the bottleneck $\obar{X}$ (and, analogously, $\obar{Y}$) can be found by minimising the functional
\begin{equation}
    \mathcal{L}[p(\obar{x}|x)] = \I(\obar{X};X) - \beta\cdot \I(\obar{X}; Y),
\end{equation}
where the Lagrange multiplier $\beta$ governs the trade-off.
Later, it has been observed that neural networks can be fruitfully analysed within this framework: considering the mutual information between the layers and the input and output variables, the training task can be seen as ``an information theoretic trade-off between compression and prediction'' \cite{Tishby2015}.
In related work, certain stochastic neural nets -- of which VAEs are a special case -- have been shown to minimise the IB functional \cite{Achille2018b}.
One can, thus, train a VAE-like stochastic neural net to learn a compression of the input and then use the encoder function without noise to construct macrovariables that supervene on the input.

\subsection{Causal Autoencoders} 
\label{ssec:case_for_NNs}

\begin{figure}[t]		
\def\layersep{3cm}

\centering
\begin{tikzpicture}[shorten >=1pt,arrows={-latex'},draw=black!50, node distance=\layersep]
    \tikzstyle{neuron}=[circle,fill=black!25,minimum size=12pt,inner sep=0pt]
    \tikzstyle{input neuron}=[neuron, fill=green!80];
    \tikzstyle{output neuron}=[neuron, fill=red!60];
    \tikzstyle{bottleneck neuron}=[neuron, fill=blue!50];
    \tikzstyle{h1 neuron}=[neuron, fill=black!30];
    \tikzstyle{h2 neuron}=[neuron, fill=black!30];
    \tikzstyle{bottleneck output neuron}=[neuron, fill=orange!50];
    \tikzstyle{annot} = [text width=4em, text centered]

    \foreach \name / \y in {1,...,5}
        \node[input neuron] (I-\name) at (0,-\y*.7 cm) {};
    \foreach \name / \y in {1,...,4}
        \path[yshift=-0.35cm]
            node[h1 neuron] (H1-\name) at (1.1*\layersep,-\y*.7 cm) {};
    \foreach \name / \y in {1,...,2}
        \path[yshift=-1.15cm]
            node[bottleneck neuron] (B-\name) at (2*\layersep,-\y*.7 cm) {};
    \foreach \name / \y in {1,...,4}
        \path[yshift=-0.55cm]
            node[h2 neuron] (H2-\name) at (2.9*\layersep,-\y*.7 cm) {};
    \foreach \name / \y in {1,...,2}
        \path[yshift=0.6cm]
            node[bottleneck output neuron] (BO-\name) at (2.6*\layersep,-\y*.7 cm) {};
    \foreach \name / \y in {1,...,5}
        \path[yshift=-0.2cm]
            node[output neuron] (O-\name) at (4*\layersep,-\y*.7 cm) {};
            
    \foreach \source in {1,...,5}
        \foreach \dest in {1,...,4}
            \path (I-\source) edge (H1-\dest);
    \foreach \source in {1,...,4}
        \foreach \dest in {1,...,2}
            \path (H1-\source) edge (B-\dest);
    \foreach \source in {1,...,2}
        \foreach \dest in {1,...,4}
            \path (B-\source) edge (H2-\dest);
    \foreach \source in {1,...,2}
        \foreach \dest in {1,...,2}
            \path (B-\source) edge (BO-\dest);
    \foreach \source in {1,...,4}
        \foreach \dest in {1,...,5}
            \path (H2-\source) edge (O-\dest);
            
    \node[annot,above of=I-1, node distance=0.7cm] (IL) {$X$};
    \node[annot,above of=B-1, node distance=0.7cm] (BL) {$\obar{X}$};
    \node[annot,above of=O-1, node distance=0.7cm] (OL) {$\hat{Y}$};
    \node[annot,above of=BO-1, node distance=0.7cm] (BOL) {$\hat{\obar{Y}}$};
\end{tikzpicture}
\caption{Structure of $\net_X$, which constitutes one half of the CAE: for input $X$, $\net_X$ learns a lower-dimensional embedding $\obar{X}$ (providing the causal macrovariables), from which it predicts $Y$ as well as the current bottleneck layer $\obar{Y}$ of $\net_Y$. The latter is the the second half of the CAE, predicting $X$ and $\obar{X}$ from $Y$, and has the same structure as $\net_X$.}\label{NN1}
\end{figure}
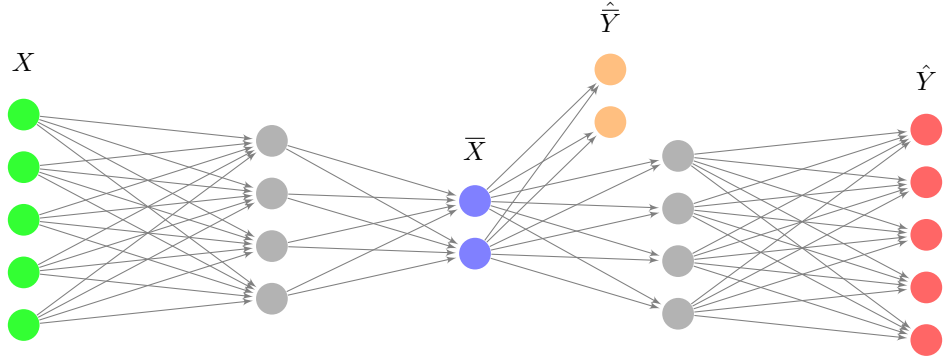

The approach presented here consists in training two stochastic neural nets simultaneously.
In the following, we will describe only the net that takes $X$ as input ($\net_X$), for simplicity of presentation;
the second net ($\net_Y$) has the same structure and together they form the Causal Autoencoder (CAE).
The most significant difference between VAEs and $\net_X$ is that the latter does not aim to decode the bottleneck layer $\obar{X}$ back into $X$; in line with the bottleneck's identification with causal macrovariables and the IB analysis discussed above, $\net_X$ is instead trained to predict $Y$ (fig.~\ref{NN1}).
In this regard, $\net_X$ is more akin to supervised learning algorithms than autoencoders.
In another deviation from VAEs, $\net_X$ has a second output layer, which is trained to predict the current bottleneck layer $\obar{Y}$ of $\net_Y$ (fig.~\ref{NN1}).
This allows, first, to ensure that the macrovariables $\obar{X}$ and $\obar{Y}$ indeed stand in some functional relation and, second, to enforce constraints on this functional relation: in the example of fig.~\ref{NN1}, the single fully connected layer between $\obar{X}$ and $\hat{\obar{Y}}$ makes the CAE learn macrovariables which stand in a linear relation to each other.
I will return to the topic of functional constraints in section \ref{ssec:constrain}.

As $\net_X$ should be trained to predict the bottleneck layer of $\net_Y$ and vice versa, both nets need to be trained simultaneously.
Taking the loss function of conventional VAEs and adding a third term for the second output layer, the loss function of $\net_X$ takes the form
\begin{equation}
\loss_{\net_X} = d_1(Y, \hat{Y}) + \beta \cdot \DKL(\mathcal{N}(0,1) || q(\obar{X}|X)) + \gamma \cdot d_2(\obar{Y}, \hat{\obar{Y}})\label{loss}
\end{equation}
where $d_1$ and $d_2$ are appropriate metrics -- like MSE -- and $q(\obar{X}|X)$ denotes the (noisy) distribution of the bottleneck neurons.
While the loss function proposed in the original VAE paper \cite{Kingma2014} did not contain any parameters, later work by \citet{Higgins2017} introduced a parameter $\beta$, as in eq.~(\ref{loss}).
Note that this parameter governs a similar trade-off as the parameter $\beta$ in the IB framework.
\citet{Higgins2017} consider it a limitation of their approach that it is not possible to estimate the optimal value of $\beta$ directly; in the present context, it allows to accommodate the objectives of different scientific goals.
In general, as discussed above, there is no unique causal structure in a given system and different representations might be better for different goals.
Hence, the possibility to find different models -- which are, for example, more or less detailed -- is actually desirable.
Intuitions on the meaning of the terms are given in appendix~\ref{app:loss_terms} and examples of how choices between models can be informed are given in the experiments section below.

A naive approach for training $\net_X$ and $\net_Y$ would be to train each of them for one minibatch in alternation.
In appendix~\ref{app:comb_loss}, we explain why it is better to instead combine the loss functions for $\net_X$ and $\net_Y$ into a sum with six terms and actually treat both parts as constituents of the same neural network, the CAE.

\subsection{Architectural constraints on function classes} 
\label{ssec:constrain}

The task of $\net_X$ is essentially to learn three functions $f$, $a$, and $\alpha$ such that $\obar{X} = f(X)$, $\hat{\obar{Y}} = a(\obar{X})$ and $\hat{Y} = \alpha(\obar{X})$.
This section concerns different constraints that can be imposed on these functions through choices about the neural network architecture.
With the architecture depicted in fig.~\ref{NN1}, the only constraint (beyond complexity constraints depending on the size of the layers) is that $a$ must be linear, given that there is no hidden layer between the bottleneck layer $\obar{X}$ and the variable output layer $\hat{\obar{Y}}$.
This implies that the function $a$ is defined by a matrix $A$, with $\hat{\obar{Y}} = A \obar{X}$.
While this constraint can be dropped by adding hidden layers or an activation function, other constraints can be put in place by altering the architecture in other respects.
In our experiments, we use the combination of two constraints:

First, $\hat{y_i} = a_i x_i + b_i$ implies that each $Y$-macrovariable $y_i$ (or bottleneck neuron of $\net_Y$) is predicted based on the value of only one corresponding $X$-macrovariable $x_i$, and the prediction must be linear.
This constraint leads to variables that fulfill two of Woodward's desiderata (section \ref{ssec:causal_macro}), namely sparse representations and strong correlations, to a very high degree.
Another advantage of this constraint is that it facilitates the investigation of causal relationships, e.g. through additive noise models (section \ref{sec:ANM}).

Second, $\hat{Y} = \Sigma_i \alpha_i(x_i)$ implies that the full $Y$-microstate is predicted as a linear combination of transformations of individual $X$-variables.
This can help to make the CAE learn variables that lend themselves to more straightforward causal interpretations.
Such a constraint in necessary to prevent the CAE from learning variables that are arbitrary recombinations of a given set of variables:
For a 'good' model with $\obar{X} := x_1, x_2$ and $\obar{Y} := y_1, y_2$ s.t. $y_1 = a_1 x_1$ and $y_2 = a_2 x_2$, the variables $\obar{X}' := x_1, x_2'$ and $\obar{Y}' := y_1, y_2'$ with $x_2' = a_1 x_1 x_2$ and $y_2' = y_1 y_2$ satisfy the same pair of equations.
The constraint proposed here is a way to select the more interpretable models.

\subsection{Investigating the causal direction with ANMs} 
\label{sec:ANM}

One advantage of the novel CAE approach is that continuous macrovariables allow to draw on an ample causal inference literature for the investigation of relationships between the detected variables \cite{Peters2017}.
In this section, we sketch the idea behind additive noise models (ANMs) before describing how they can be adapted to overcome difficulties faced in the present context.

\subsubsection{Additive Noise Models} 

In the general framework of structural causal models, a relationship of $x_i$ causing $y_i$ can be represented by some assignment $y_i := \hat{\alpha_i}(x_i, n)$, with the noise $n$ being independent from $x_i$.
Now the idea behind ANMs is that by imposing plausible restrictions on $\hat{\alpha_i}$, it can be possible to infer causal relationships from observational data.
As suggested by their name, ANMs assume that the independent noise is additive, i.e. we can reformulate the assignment as $y_i := \alpha_i(x_i) +n$, where $x_i$ and $n$ are again independent.
Given two dependent variables $x_i$ and $y_i$, we can try to find such an ANM either for the causal direction $x_i \rightarrow y_i$ or the reverse $x_i \leftarrow y_i$.
\citet{Mooij2016} compare the performance of some approaches to this causal inference task on a benchmark dataset and come to the conclusion that ``the original ANM method ANM-pHSIC proposed by Hoyer et al. (2009) turned out to be one of the best methods overall'' (p. 45).
On the mentioned approach, a prediction $\hat{y}_i$ of $y_i$ is made based on the predictor $x_i$, in order to compute the residual $y_{i,res} = y_i - \hat{y_i}$ that serves as a proxy for the noise $n$.
Then it is tested whether $y_{i,res}$ and $x_i$ are independent; for this, Hoyer and colleagues suggest to use the Hilbert-Schmidt Independence Criterion (HSIC, \cite{Gretton2005}).
The same steps are applied with $x_i$ and $y_i$ reversed; if the independence hypothesis (and thus the ANM) is accepted in one direction but rejected in the other, we infer that this is the causal direction.

\subsubsection{CAEs and ANMs} 
\label{ssec:ANMvsNN}

In the causal inference literature, it is usually assumed that the causal variables are given and come with a natural scale.
However, the correct numerical representation of causal variables is often not clear in complex applications like neuro- or climate science.
Here, the causal patterns we want to investigate might not have a privileged numerical representation and it might not be obvious whether some macrovariable $x_i$ is better than e.g. a transformed version $log(x_i)$.
This immediately leads to a problem for directly investigating the macrovariables detected by a CAE through ANMs.

Even with the architectural constraints described in section \ref{ssec:constrain}, the CAE is still agnostic with respect to monotonic transformations of the detected variables.
Therefore, the CAE should not be seen as learning macrovariable pairs but equivalence classes of macrovariable pairs, where pairs are equivalent if their variables can be transformed into each other by monotonic transformations.
This is not surprising from an information-theoretic perspective, as mutual information is invariant under invertible (and, thus, monotononic) transformations.
Now the issue for applying ANMs is that the independence of residuals is affected by such transformations.
Therefore, the ANM approach can yield very different results for two pairs of variables even if both pairs are equivalent from the CAE's perspective.
This means that we cannot naively plug the detected macrovariables directly into ANM algorithms for causal inference.

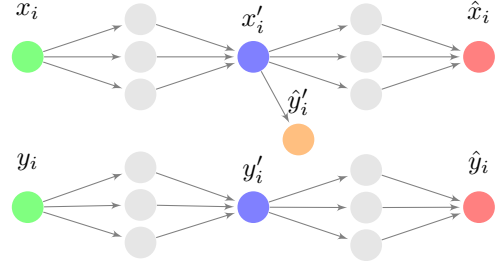
\begin{wrapfigure}{r}{.5\textwidth}	
\def\layersep{1.5cm}
\centering
\begin{tikzpicture}[shorten >=1pt,arrows={-latex'},draw=black!50, node distance=\layersep]
    \tikzstyle{neuron}=[circle,fill=black!25,minimum size=12pt,inner sep=0pt]
    \tikzstyle{input neuron}=[neuron, fill=green!50];
    \tikzstyle{output neuron}=[neuron, fill=red!50];
    \tikzstyle{bottleneck neuron}=[neuron, fill=blue!50];
    \tikzstyle{bottleneck output neuron}=[neuron, fill=orange!50];
    \tikzstyle{h1 neuron}=[neuron, fill=black!10];
    \tikzstyle{h2 neuron}=[neuron, fill=black!10];
    \tikzstyle{annot} = [text width=4em, text centered]

    \node[input neuron] (I-1) at (0,-1) {};
    \node[input neuron] (I-2) at (0,-3) {};
    \foreach \name / \y in {1,2,3}
        \path[yshift=0cm]
            node[h1 neuron] (H1-\name) at (1*\layersep,-\y*0.5cm) {};
    \foreach \name / \y in {4,5,6}
        \path[yshift=-0.4cm]
            node[h1 neuron] (H1-\name) at (1*\layersep,-2-\y*0.5cm) {};
    \path[yshift=0cm]
         node[bottleneck neuron] (B-1) at (2*\layersep,-1cm) {};
    \path[yshift=0cm]
         node[bottleneck neuron] (B-2) at (2*\layersep,-3cm) {};
    \path[yshift=0cm]
         node[bottleneck output neuron] (BO) at (2.4*\layersep,-2.1cm) {};
    \foreach \name / \y in {1,2,3}
        \path[yshift=0cm]
            node[h2 neuron] (H2-\name) at (3*\layersep,-\y*0.5cm) {};
    \foreach \name / \y in {4,5,6}
        \path[yshift=-0.4cm]
            node[h2 neuron] (H2-\name) at (3*\layersep,-3-\y*0.5cm) {};
    \path[yshift=0cm]
         node[output neuron] (O-1) at (4*\layersep,-1cm) {};
    \path[yshift=0cm]
         node[output neuron] (O-2) at (4*\layersep,-3cm) {};
            
     \foreach \dest in {1,2,3}
         \path (I-1) edge (H1-\dest);
     \foreach \dest in {4,5,6}
         \path (I-2) edge (H1-\dest);
    \foreach \source in {1,2,3}
         \path (H1-\source) edge (B-1);
    \foreach \source in {4,5,6}
         \path (H1-\source) edge (B-2);
     \foreach \dest in {1,2,3}
         \path (B-1) edge (H2-\dest);
     \path (B-1) edge (BO);
     \foreach \dest in {4,5,6}
         \path (B-2) edge (H2-\dest);
    \foreach \source in {1,2,3}
         \path (H2-\source) edge (O-1);
    \foreach \source in {4,5,6}
         \path (H2-\source) edge (O-2);
            
    \node[annot,above of=I-1, node distance=0.6cm] (IL) {$x_i$};
    \node[annot,above of=B-1, node distance=0.5cm] (BL) {$x_i'$};
    \node[annot,above of=O-1, node distance=0.6cm] (OL) {$\hat{x}_i$};
    \node[annot,above of=I-2, node distance=0.6cm] (IL) {$y_i$};
    \node[annot,above of=B-2, node distance=0.5cm] (BL) {$y_i'$};
    \node[annot,above of=O-2, node distance=0.6cm] (OL) {$\hat{y}_i$};
    \node[annot,above of=BO, node distance=0.5cm] (BO) {$\hat{y}_i'$};
\end{tikzpicture}
\caption{A pair of VAEs is used to find the monotonically transformed variables $x_i'$ and $y_i'$ (from $x_i$ and $y_i$) that minimise the dependence (given by the HSIC score) between $x_i'$ and $y_{i,res}' = y_i' - \hat{y}_i'$.}\label{VAE_Hoyer}
\end{wrapfigure}
In order to check whether there are transformations of detected variables $x_i$ and $y_i$ that are compatible with an ANM in one of the two directions, we can check the transformations which minimise the dependence between residual and predictor.
The first task is, thus, to find two variable pairs satisfying $x_i', y_i' = \arg \underset{x_i, y_i}{\min}\ \HSIC(x_i, y_{i,res})$ and $x_i'', y_i'' = \arg \underset{x_i, y_i}{\min}\ \HSIC(y_i, x_{i,res})$, where all variables are monotonic transformations of $x_i$ and $y_i$, respectively.
One way to do this is to again use VAEs, one for each variable (fig.~\ref{VAE_Hoyer}).
To find the variables that minimise the dependence between $x_i$ and $y_{i,res}$, the VAEs are trained on the loss
\begin{equation}
\loss_{X \rightarrow Y} = \lossVAEx + \lossVAEy + \frac{\mse(y_i, \hat{y}_i)}{\var(y_i)} + \HSIC(x_i, y_{i, res}).
\end{equation}

After finding the optimal transformations, the HSIC score of $x_i'$ and $y_{i,res}'$ is computed.\footnote{The idea of minimising the HSIC score directly before testing it has already been explored in \cite{Mooij2009}, although for regression and not in the setting of an autoencoder.}
The same procedure is applied with $x_i$ and $y_i$ reversed, to find a monotonically transformed variable pair with minimal dependence between $y_i''$ and $x_{i,res}''$.
The two scores are then compared to see whether there is a strong case for a causal influence in either direction.
A threshold can be computed that gives a criterion deciding whether to accept or reject the independence hypothesis \cite{Gretton2009}.
However, as whether the threshold is exceeded depends on both the sample size and the details of the loss function, it seems advisable to also take into account the disparity between $\HSIC(x_2', y_{2,res}')$ and $\HSIC(y_2'', x_{2,res}'')$.

\section{Experiments} 
\label{sec:exp1}

\subsection{Simulated data} 
\label{ssec:detec_synth}

We first report experiments on simulated data with a known ground truth model.\footnote{The code for all experiments is available at \url{https://github.com/benedikthoeltgen/causal-macro}.}
Here, $X$ and $Y$ are random variables over $\RR^{64}$, so they can be thought of as quadratic grey-scale images of $8 \times 8$ pixels.
The underlying ground truth model used for generating the data is that of fig.~\ref{two_macrovar}:
There are four macrovariables $x_1, y_1, x_2, y_2$ where $x_1$ and $y_1$ have a common cause $c_1$ and $y_2$ is caused by $x_2$.
The four macrovariables correspond to averages in the left/right half of $X$ and the top/bottom half of $Y$, respectively.
The data is generated according to the following three structural assignments:
\begin{equation*}
x_1 := c_1 + n^X_1\quad\quad\quad\quad\quad\quad
y_1 := c_1^3 + n^Y_1\quad\quad\quad\quad\quad\quad
y_2 := \tanh(x_2)+ n^Y_2
\end{equation*}
with $c_1, x_2 \sim \mathcal{U}([-1,1])$ and  $n^X_1, n^Y_1, n^Y_2 \sim \mathcal{U}([-0.2,0.2])$ all uniformly distributed and mutually independent.
Pairs of low-level states $(x,y)$ are generated by starting with states that satisfy the high-level descriptions exactly and then adding pixel-wise uniform noise in $[-0.2,0.2]$.

\subsubsection{Variable detection}

We use the constrained CAE structure discussed above with bottleneck dimension 4.
The results of different hyperparameter settings are reported in table~\ref{table:synth}.
It shows that both $\net_X$ and $\net_Y$ effectively use two bottleneck neurons for all hyperparameter settings (i.e. the other neurons are dominated by the injected noise and cannot carry information), corresponding to the number of macrovariables in the ground truth model (see previous paragraph), attesting the robustness of the approach when there is a clear ground truth.
However, the CAE's ability to learn variables that can predict each other (e.g. $y_1$ from $x_1$ and vice versa) varies greatly.
Although the network structure generally yields pairs of variables $x_i$ and $y_i$ that predict each other, this is not the case for low choices of $\gamma$; this sometimes leads to negative explained variance (EV) scores in predicting the variables.
The CAE with the best performance according to the EV measures is the one with the highest $\gamma$ and lowest $\beta$, i.e. the one where the relative weight of the second term of the loss function is the lowest.
This is not surprising, as it means that the training is less noisy (no noise was injected during the evaluation on the validation set) and, thus, more accurate -- yet still generalisable -- predictions can be learned.
\begin{center}
\captionof{table}{Explained variance (EV) in predicting $Y$/$X$ and $\obar{Y}/\obar{X}$ through $\net_X/\net_Y$ on the test set and the number of detected variables $|\obar{X}|,\ |\obar{Y}|$ (i.e. number of informative/non-random bottleneck neurons) after 1500 epochs, for different values of $\beta$ and $\gamma$ in eq.~(\ref{loss}).
The number of detected macrovariables is always two, corresponding to the number of ground-truth variables.
For some settings with low $\gamma$, the learned variables do not allow a prediction of their counterparts, resulting in negative EV scores.}\label{table:synth}
\begin{tabular}{L @{\kern.7cm} C @{\kern.7cm} C @{\kern.7cm} C}
\toprule
 & \beta=1 & \beta=0.1 & \beta=0.01 \\
\midrule
\gamma=1 & |\obar{X}|=2,\ |\obar{Y}|=2 & |\obar{X}|=2,\ |\obar{Y}|=2 & |\obar{X}|=2,\ |\obar{Y}|=2 \\
& EV(Y/X)=.62/.57; & EV(Y/X)=.80/.77; & EV(Y/X)=.81/.78; \\
& EV(\obar{Y}/\obar{X})=.62/.63; & EV(\obar{Y}/\obar{X})=.89/.88; & EV(\obar{Y}/\obar{X})=.89/.90; \\
\midrule
\gamma=0.1 & |\obar{X}|=2,\ |\obar{Y}|=2 & |\obar{X}|=2,\ |\obar{Y}|=2 & |\obar{X}|=2,\ |\obar{Y}|=2 \\
& EV(Y/X)=.57/.51; & EV(Y/X)=.80/.77; & EV(Y/X)=.81/.78; \\
& EV(\obar{Y}/\obar{X})={<}0/{<}0; & EV(\obar{Y}/\obar{X})=.86/.86; & EV(\obar{Y}/\obar{X})=.89/.87; \\
\midrule
\gamma=0.01 & |\obar{X}|=2,\ |\obar{Y}|=2 & |\obar{X}|=2,\ |\obar{Y}|=2 & |\obar{X}|=2,\ |\obar{Y}|=2 \\
& EV(Y/X)=.59/.53; & EV(Y/X)=.80/.77; & EV(Y/X)=.81/.78; \\
& EV(\obar{Y}/\obar{X})={<}0/{<}0; & EV(\obar{Y}/\obar{X})={<}0/{<}0; & EV(\obar{Y}/\obar{X})=.61/.26; \\
\bottomrule
\end{tabular}
\end{center}

\subsubsection{Causal direction}

For analysing the causal direction, the variables detected by the CAE with hyperparameters $\beta=0.01$ and $\gamma=1$ were chosen since this setting showed the best performance w.r.t. the explained variance metrics (cf. table \ref{table:synth}).
Recall that to satisfy an ANM, the residual in predicting one variable should be independent from the respective predictor variable.
After applying the respective transformations, the HSIC scores for the second pair differ by a factor of ten, as $\HSIC(x_2', y_{2,res}') = 0.32$ and $\HSIC(y_2'', x_{2,res}'') = 3.15$.
In particular, the former drops below the threshold while the latter remains above it.
This implies that the ANM $y_2' = \alpha_2(x_2') + n_2^Y$ is accepted while the reverse model $x_2'' = \alpha_2(y_2'') + n_2^X$ is rejected.
By the criterion suggested in \cite{Hoyer2009}, we correctly infer that $x_2$ causes $y_2$.

For the first variable pair $(x_1, y_1)$, neither variable causes the other in the generative model.
After transforming the detected variables, we get $\HSIC(x_2', y_{2,res}') = 3.88$ and $\HSIC(y_2'', x_{2,res}'') = 13.92$.
All scores are well above the threshold, so we infer that no variable causes the other, in line with the ground truth.
It should also be noted that one value is about four times as high as the other, both before and after the transformation.
This is less salient than the disparity for the causal variables $x_2$ and $y_2$; still, it shows that the present approach might run into problems for more complex datasets.

\subsection{Natural data: El Ni\~{n}o} 

Here, we report results from running the algorithm on the climate dataset investigated in \cite{Chalupka2016b}, retrieved from the first author's website.
It comprises 13140 weekly averaged measurements of zonal winds (ZW) and sea surface temperatures (SST) in the equatorial pacific, each on a $9 \times 55$ grid spanning from 140°E to 80°W and 10°N to 10°S, from the years 1979-2014.
A well-known climate phenomenon repeatedly appearing in this region is El Ni\~{n}o.
According to the National Oceanic and Atmospheric Administration (NOAA), it is defined as a ``three-month average of sea surface temperature departures from normal for a critical region of the equatorial Pacific'' \cite{NOAA}.
A good reason to study El Ni\~{n}o in the context of causal macrovariable detection is that it provides strong causal links between various measurable quantities (such as ZW and SST) which allow both a high- and a low-level description.
\citet{Chalupka2016b} apply their causal feature learning algorithm to this dataset and interpret their results as an unsupervised discovery of the El Ni\~{n}o phenomenon.

The CAE again uses the constrained structure, now with 16 bottleneck neurons.
On this dataset, different choices of hyperparameters not only lead to differences in the accuracy of predictions but also in the number of detected variables (table \ref{table:nino}).
Low choices of $\beta$ -- and, to a smaller degree, low choices of $\gamma$ -- led to higher numbers of variables, i.e. non-noise bottleneck neurons.
This higher model complexity comes with higher predictive power, as discussed in appendix~\ref{app:loss_terms}. 
Low choices of $\gamma$ again lead to negative EV scores in predicting the variables.
One can also see that $|\obar{Y}|$ is often greater than $|\obar{X}|$.
This is probably because $X$ is harder to predict from $Y$ than vice versa (see the EV scores).

\begin{center}
\captionof{table}{Explained variance (EV) in predicting $Y$/$X$ and $\obar{Y}/\obar{X}$ through $\net_X/\net_Y$ on the test set and the number of detected variables $|\obar{X}|,\ |\obar{Y}|$ for different values of $\beta$ and $\gamma$ in eq.~(\ref{loss}).
Here, the number of detected variables strongly depends on the choice of hyperparameters, especially $\beta$.
For some settings with low $\gamma$, no prediction of the variables is learned, resulting in negative EV scores.}\label{table:nino}
\begin{tabular}{L @{\kern.7cm} C @{\kern.7cm} C @{\kern.7cm} C}
\toprule
 & \beta=1 & \beta=0.1 & \beta=0.01 \\
\midrule
\gamma=1 & |\obar{X}|=3,\ |\obar{Y}|=2 & |\obar{X}|=4,\ |\obar{Y}|=5 & |\obar{X}|=7,\ |\obar{Y}|=7 \\
& EV(Y/X)=.70/{<}0; & EV(Y/X)=.83/.52; & EV(Y/X)=.86/.64; \\
& EV(\obar{Y}/\obar{X})=.82/.43; & EV(\obar{Y}/\obar{X})=.82/.85; & EV(\obar{Y}/\obar{X})=.88/.88; \\
\midrule
\gamma=0.1 & |\obar{X}|=3,\ |\obar{Y}|=3 & |\obar{X}|=6,\ |\obar{Y}|=9 & |\obar{X}|=13,\ |\obar{Y}|=13 \\
& EV(Y/X)=.71/.15; & EV(Y/X)=.85/.67; & EV(Y/X)=.90/.67; \\
& EV(\obar{Y}/\obar{X})={<}0/{<}0; & EV(\obar{Y}/\obar{X})={<}0/.41; & EV(\obar{Y}/\obar{X})=.46/.46; \\
\midrule
\gamma=0.01 & |\obar{X}|=3,\ |\obar{Y}|=4 & |\obar{X}|=7,\ |\obar{Y}|=10 & |\obar{X}|=16,\ |\obar{Y}|=16 \\
& EV(Y/X)=.71/.21; & EV(Y/X)=.86/.69; & EV(Y/X)=.90/.78; \\
& EV(\obar{Y}/\obar{X})= {<}0/{<}0; & EV(\obar{Y}/\obar{X})={<}0/{<}0; & EV(\obar{Y}/\obar{X})={<}0/{<}0; \\
\bottomrule
\end{tabular}
\end{center}

\begin{figure}
\begin{center}
\includegraphics[width=.8\textwidth]{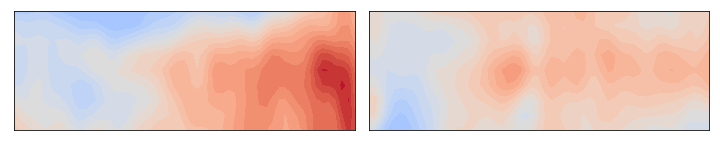}
\end{center}
\caption{Deviation from the mean temperature for the inputs with the highest values for two of the five learned macrovariables over the temperature dataset $Y$ (for $\beta=0.1, \gamma=1$). For all settings of $\beta$ and $\gamma$, the CAE detects two variables that allow to distinguish between the eastern 'warm pool' (WP, left) and the western/central 'cold tongue' (CT, right) El Ni\~{n}o variations (see appendix~\ref{app:nino_kug}).}\label{fig:nino}
\end{figure}

The detected variables allow a more nuanced analysis than the clusters in \cite{Chalupka2016b} as they not only represent a warm area in general, but also allow to distinguish between two different variations of El Ni\~{n}o:
For any tested configuration of $\beta$ and $\gamma$, there is one neuron tracking a very warm tongue from the east and one tracking a warm area in the western centre (fig.~\ref{fig:nino}).
Remarkably, these patterns correspond to the two known variations of El Ni\~{n}o, the warm pool (WP) and cold tongue (CT) El Ni\~{n}o \cite{Kug2009}.
As shown in appendix~\ref{app:nino_kug}, ``the warm events can be well separated into the WP El  Ni\~{n}o and CT El Ni\~{n}o based on their SST anomaly patterns''.
Note that while the macrovariables assign a value to the strength of some pattern \emph{for each week}, the labels CT, WP, or mixed are assigned \emph{for each year}.
As El Ni\~{n}o is defined as an anomaly over several months, assigning macrovariables based on potential causal relationships on a weekly scale (as done both here and by \cite{Chalupka2016b}) appear problematic.
For these reasons, we did not expect to find causal relationships between macrovariables through the ANM approach in the climate dataset.
And indeed, the ANM analysis yielded no evidence for a direct causal relationship (see appendix~\ref{app:nino_corr}).

\section{Discussion} 

As causal macrovariables form information bottlenecks, they can be captured by bottleneck layers of a novel neural network called Causal Autoencoder (CAE).
Different hyperparameter settings in the loss function allow to weight the model's predictive power, simplicity, and accuracy against each other.
In experiments, the CAE recovers the ground-truth variables from data generated by a simple four variable model.
On natural data, it detects sensible variables that align with known variations of the El Ni\~{n}o phenomenon.
It is also possible to apply additive noise models to the detected variables after a transformation step and thus investigate the causal relationship.
For the simulated data, this has been shown to correctly identify the causal relationships.
It is, however, not yet clear whether the application of ANMs will prove to be useful in scientific practice.
In an experiment performed on natural climate data, the ANM approach did not lead to additional insights.
It is not clear whether this is due to particularities of the used data.
Another worry relates to the general difficulty of inferring causal relationships from observational data; Woodward's first criterion for causal variables cited in section \ref{ssec:causal_macro} demands that interventions on these variables have distinctive effects.
While the other three cited criteria are reasonably well met, this one is hard to assess.

When two variables both stand in a direct causal relationship and share a common cause, the ANM procedure can be expected to fail.
In such cases, other approaches like the Neural Causation Coefficient \cite{Lopez-Paz2017} might give better results.
On this issue, further theoretical and empirical investigations are required.
Another avenue for future research is the investigation of variations of the CAE, in particular of other architectural constraints.
Another exciting possibility would be to build CAEs in the form of a recurrent (RNN) or convolutional neural net (CNN).
CAE-RNNs might be better suited for the application to time-series data, such as the climate data used in this work.
Extending the CAE to CNNs could be particularly useful for domains where the location of a pattern is not important, i.e. where the same causal phenomenon can appear in different parts of a dataset.
Lastly, it would be very interesting to investigate potential theoretical connections between the mutual information in observational distributions, as used for discovery here, and information-theoretic measures of causality.
A variety of such measures can be found in the literature, often employing the mutual information between interventional distributions \cite{Ay2008, Griffiths2015, Hoel2017}.
This might provide a more thorough theoretical foundation of the approach proposed in this work.

\section*{Acknowledgements}

I would like to thank Stephan Hartmann, Moritz Grosse-Wentrup, Frederick Eberhardt, Gunnar König, Timo Freiesleben, Lood van Niekerk, and three reviewers for helpful discussions and thorough feedback at different stages of this project.


\newpage
\bibliography{library}
\normalsize

\newpage
\appendix

\section{The CAE loss function}

\subsection{Loss terms and trade-offs}
\label{app:loss_terms}

To demonstrate the benefits of the freedom to choose the parameters, we can give a high-level description of the role that each of the terms in the loss function (equation~\ref{loss})
\begin{equation*}
\loss_{\net_X} = d_1(Y, \hat{Y}) + \beta \cdot \DKL(\mathcal{N}(0,1) || q(\obar{X}|X)) + \gamma \cdot d_2(\obar{Y}, \hat{\obar{Y}})
\end{equation*}
plays for the resulting causal macrovariable model.

The \emph{first term} simply measures the distance (e.g. via MSE) between the prediction $\hat{Y}$ and the known $Y$ (for $\net_X$).
A low loss in this first term signifies that the bottleneck neurons, and hence the macrovariables, contain much of the information necessary for predicting the output.
In other words, the resulting model has a high predictive power as it captures large parts of the system in question.

The \emph{second term} measures the Kullback-Leibler Divergence of the noisy bottleneck distribution from the standard normal distribution.
$\DKL$ is zero for a single neuron if its activation is always drawn completely at random from that distribution and thus carries no information about $X$.
It increases when the activation is drawn from other normal distributions, in particular when the mean varies among samples and the noise is small, which allows the neuron to carry information.
The more neurons carry information, and the more information they carry, the higher is the loss.
I denote by $|\obar{X}|$ and $|\obar{Y}|$ the number of neurons that carry information, which can be lower than the number of neurons in the bottleneck.
A low loss in this second term signifies that the model is fairly simple (comprising few variables) and fairly robust, as it has been trained on noisy variables.

The \emph{third term} is introduced in this work for the novel bottleneck neuron output layer. 
Similar to the first term, it measures (for $\net_X$) the distance between the prediction $\hat{\obar{Y}}$ and $\obar{Y}$, the latter being calculated from the current $\net_Y$.
A low loss in this third term signifies that the $X$-variables allow a good prediction of the $Y$-variables, implying that the model is fairly accurate and may have closely causally connected variables.

\subsection{Combining the loss functions}
\label{app:comb_loss}

An issue that comes up especially -- though not exclusively -- for cases of asymmetric information (i.e. when $X$ allows a better prediction of $Y$ than the other way round or vice versa)\footnote{Note that the formulation 'asymmetric information' might be misleading as mutual information is symmetric, that is, $\I(X;Y) = \I(Y;X)$. What is asymmetric, then, is how much information each of the variables contain \emph{in addition} to the information shared between the two.}, is the coordination between $\net_X$ and $\net_Y$.
Such cases motivate the use of a combined loss function $\loss_{\net_X} + \loss_{\net_Y}$ instead of training the two nets in alternation.
This is best illustrated through an example.

Consider a system where $y_1$ and $x_1$ are given by the average of $Y$ and $X$, respectively, and where $x_1  \sim \mathcal{U}([-1,1])$ and $y_1 = x_1^2$.
Here, for each $Y$-sample with $y_1=q$ for some value $q \in [0,1]$, the corresponding $X$-sample has a $x_1$-value of either $\sqrt{q}$ or $-\sqrt{q}$ with equal probability.
So $\net_Y$ has no chance of learning a useful prediction of $x_1$ or $X$; in fact, the third term of its loss function is minimal if it outputs the prediction $\hat{x}_1 = 0$ for every sample (given a loss function like MSE).
As detecting $y_1$ does not help $\net_Y$ to decrease its loss, it would come up with a completely noisy bottleneck layer.
This keeps the CAE from finding a model with useful variables like $x_1$ and $y_1$, as it relies on $\net_Y$ to detect $y_1$.
It is possible to overcome this problem by using a combined loss function for gradient descent, by treating $\net_X$ and $\net_Y$ as \emph{one} neural network.
This way, both parts of the CAE can adapt also in order to reduce each other's loss.
In the mentioned example, $\net_X$ could learn $x_1' = x_1^2$ -- which is sufficient to predict both $y_1$ and $Y$ --, such that $\net_Y$ would be able to reduce its third loss term by learning $y_1$ and thereby predicting $x_1' = y_1$.
$\net_X$, on the other hand, can now predict $y_1$, and thus $\obar{Y}$ in general, more accurately, and reduce its third loss term.
Whether such an optimal solution will be found eventually still depends on the learning process and on local minima in particular.
But as the huge successes of neural networks have shown, this issue often turns out to be less grave than expected.
As this simple example shows, a combined loss function can help the CAE to find better models for a given specification of hyperparameters.

\newpage

\section{Transformations of detected variables for simulated data}
\label{app:simu_res}

Here, we give more details on the transformation of the variables and the  application of ANMs in the simulated data experiments.
As fig.~\ref{fig:trans_scatter} (a) indicates, $y_{2,res}$ and $x_2$ (left side) appear to be less dependent than $x_{2,res}$ and $y_2$ (right side), in line with the true causal direction $x_2 \rightarrow y_2$.
This impression is partly confirmed by calculating the HSIC scores (as suggested by \citet{Hoyer2009}):
The test statistics are $\HSIC(x_2, y_{2,res}) = 5.18$ and $\HSIC(y_2, x_{2,res}) = 13.76$ respectively, with a threshold of 0.65.
This difference is, arguably, not sufficient for accepting the causal model $x_2 \rightarrow y_2$ (especially as both HSIC scores are far above the threshold).
This suggests that it is necessary to first apply the VAE-based transformation step to minimise dependence as described in section~\ref{ssec:ANMvsNN}.

Naturally, this yields variables with lower HSIC test statistics (see also fig.~\ref{fig:trans_scatter} (b), (c)).
They now differ by a factor of ten, as $\HSIC(x_2', y_{2,res}') = 0.32$ and $\HSIC(y_2'', x_{2,res}'') = 3.15$.
In particular, the former drops below the threshold while the latter remains above it.
This means that the ANM $y_2' = \alpha_2(x_2') + n_2^Y$ is accepted while the reverse model $x_2'' = \alpha_2(y_2'') + n_2^X$ is rejected.
By the criterion suggested in \citep{Hoyer2009}, we infer that $x_2$ causes $y_2$; this is in line with the generative model, i.e. with the known ground truth.

\begin{center}
\begin{tikzpicture}
\node (pre) {\includegraphics[width=.48\textwidth]{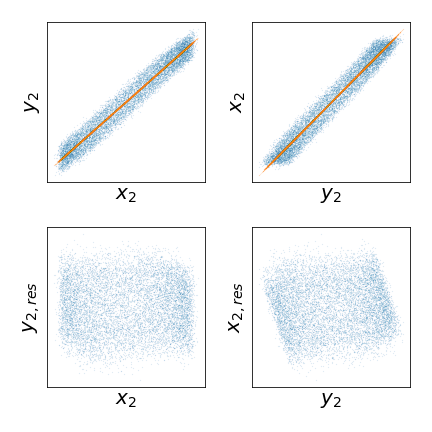}};
\node (post-x) at (-5.6, -1) {\includegraphics[width=.23\textwidth]{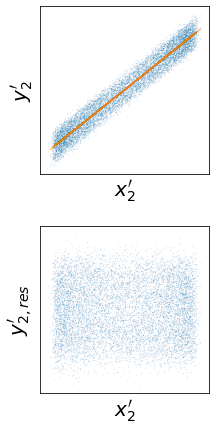}};
\node (post-y) at (5.7, -1) {\includegraphics[width=.23\textwidth]{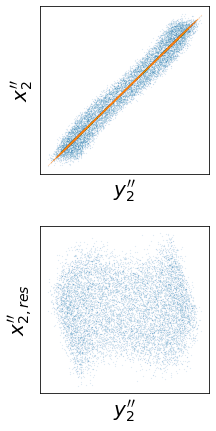}};

\node (pre-left) at (-2.8, 0.3) {};
\node (post-x-right) at (-4.2, -1) {};
\node (pre-right) at (3.3, 0.3) {};
\node (post-y-left) at (4.7, -1) {};

\draw [ultra thick,magenta,->] (pre-left) to (post-x-right);
\draw [ultra thick,magenta,->] (pre-right) to (post-y-left);

\node (a) at (0, 3.7) {(a)};
\node (b) at (-5.6, 2.6) {(b)};
\node (c) at (5.7, 2.6) {(c)};
\end{tikzpicture}
\captionof{figure}{Values and residuals of the variables initially detected by the CAE \textbf{(a)}, as well as of the transformed variables, after minimising either $\HSIC(x_2, y_{2,res})$ \textbf{(b)} or $\HSIC(y_2, x_{2,res})$ \textbf{(c)}.
Top row: Scatter plots of variable values (blue) and their predictions (orange).
Bottom row: Scatter plots of the residuals (actual value minus prediction) against the variable on which the prediction is based.
While transformations with independent $x_2'$ and $y_{2,res}'$ can be found \textbf{(b)}, the same does not hold true for $y_2''$ and $x_{2,res}''$ \textbf{(c)}.
This indicates that $x_2$ causes $y_2$, in line with the known ground-truth model.}
\label{fig:trans_scatter}
\end{center}

\newpage

\section{El Ni\~no} 
\label{app:nino_kug}

\begin{center}
    \includegraphics[width=.7\textwidth]{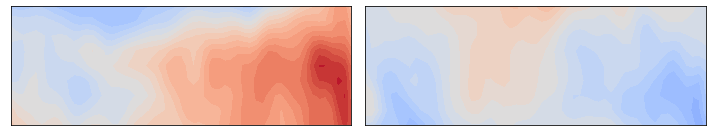}
    \includegraphics[width=.7\textwidth]{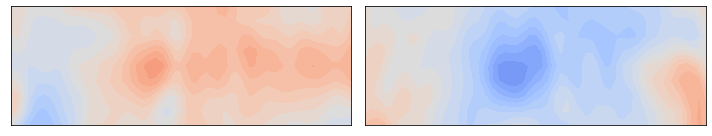}
\captionof{figure}{Deviation from mean temperature of inputs with highest (left) and lowest (right) values for two of the detected macrovariables (cf. fig.~\ref{fig:nino}).
The top row variable tracks the temperature in the east (with high values corresponding to WP El Ni\~{n}o) while the bottom row variable tracks the temperature in the center-west (with high values corresponding to CT El Ni\~{n}o), compare fig.~\ref{Kug} below.}\label{nino_high-low}
\end{center}
\vspace{20pt}

\begin{center}
    \includegraphics[width=.95\textwidth]{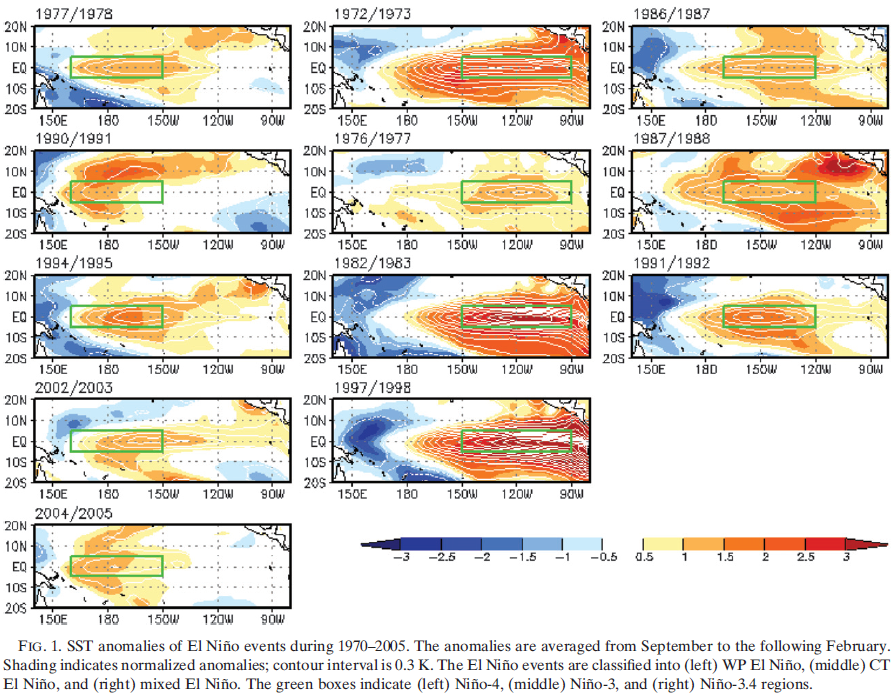}
\captionof{figure}{Variations of El Ni\~{n}o, taken from \citet[1501]{Kug2009}.
Note that the depicted segment of the Pacific Ocean is the same as in the dataset investigated in the experiments in east-west expanse while it is greater in its north-south expanse: the samples in the investigated dataset only extend from 10°N to 10°S, both of which are marked here in the sub-figures by horizontal dotted lines.
Also note that this figure shows averages over several months rather than a single week.}\label{Kug}
\end{center}

\newpage

\section{Relationships between detected climate variables} 
\label{app:nino_corr}

To investigate causal relationships between variables in the climate dataset, we looked at the variables detected by the CAE with $\beta=0.1, \gamma=1$.
This choice was based on the model's good predictive power through only a few variables (table \ref{table:nino}).
Three variable pairs predict each other, while the other three non-noise bottleneck neurons are apparently merely used by the CAE to predict the high-dimensional samples $X$ and $Y$ (see fig \ref{fig:4x5}).
Among the three pairs, no direct causal relationship can be inferred from the ANMs:
Even after the transformation step, all HSIC scores are around 1 or 2, thus showing no causal asymmetry.

\begin{center}
\centering
\includegraphics[width=.9\textwidth]{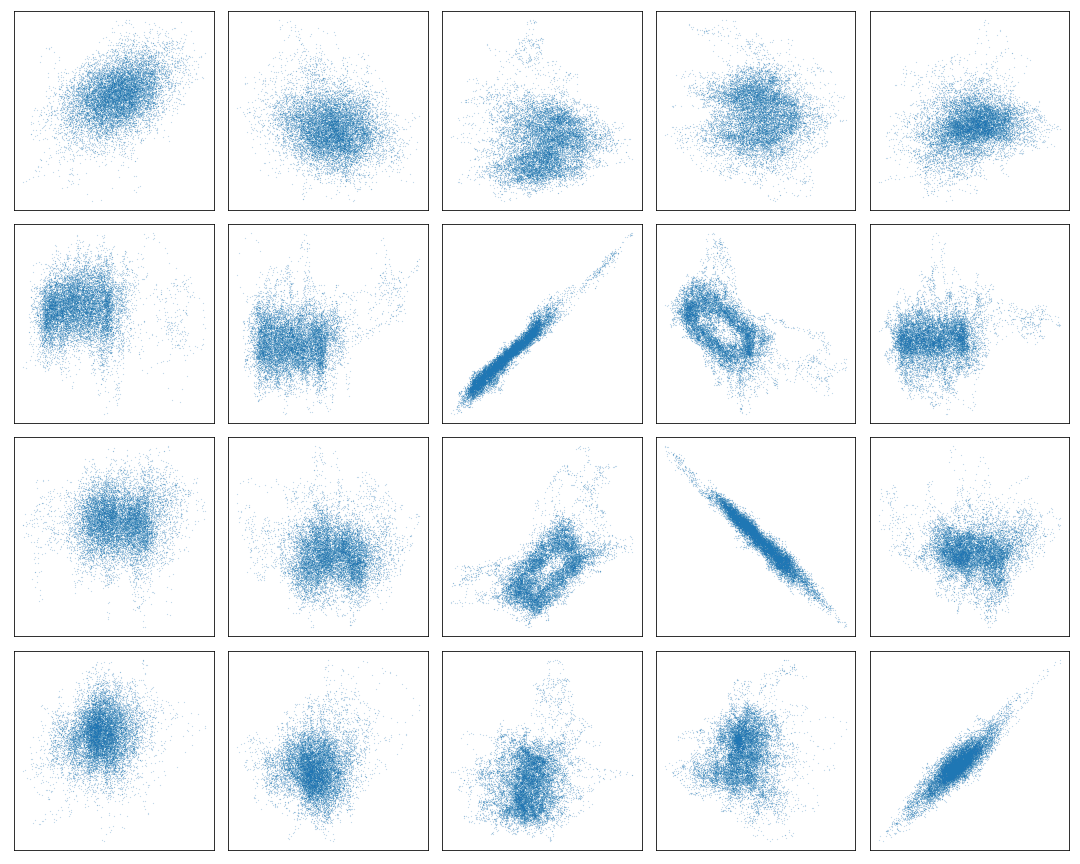}
\captionof{figure}{Scatter plots of the five $Y$-variables against the four $X$-variables detected by the CAE with $\beta=0.1, \gamma=1$ applied to the El Ni\~{n}o dataset. Three pairs of variables show high correlation, suggesting a possible causal relationship. There is, however, no clear indication about the direction of causality based on ANMs -- this visual impression is confirmed by tests. For this dataset, common causes and/or a cyclic relationship are likely.}
\label{fig:4x5}
\end{center}

\end{document}